\documentclass[letterpaper, 10 pt, conference]{ieeeconf}  

\IEEEoverridecommandlockouts                              

\newcommand{\RNum}[1]{\uppercase\expandafter{\romannumeral #1\relax}}

\usepackage{graphics} 
\usepackage{graphicx}
\graphicspath{{./figure/}}
\usepackage{subcaption}
\usepackage[font=small]{caption}

\usepackage{bm}

\usepackage[noadjust]{cite}

\usepackage{hyperref}
\usepackage{amsmath} 
\usepackage{amssymb}  
\usepackage{algorithm}
\usepackage{algorithmic}
\usepackage{romannum}
\usepackage{booktabs}
\usepackage{color}
\usepackage{xcolor}
\usepackage{enumerate}
\usepackage{svg}

\title{\LARGE \bf
Viko: An Adaptive Gecko Gripper with Vision-based Tactile Sensor
}

\author{Chohei Pang$^{1}$, Kinwing Mak$^{1}$, Yazhan Zhang$^{1}$, Yang Yang$^{2}$, Yu Alexander Tse$^{1}$ \\and Michael Yu Wang$^{3}$, \textit{Fellow, IEEE}
	\thanks{*Research is supported by the Hong Kong Innovation and Technology
Fund (ITF) ITS-104-19FP and also supported by the National Natural Science Foundation of China (Grant No. 52005269). }
	\thanks{$^{1}$C. Pang, K. Mak, Y. Zhang and Y. Tse (corresponding author) are with the Department of Mechanical and Aerospace Engineering, Hong Kong University of Science and Technology, Hong Kong (e-mail: chpangad@connect.ust.hk; kwmakaf@connect.ust.hk; yatse@connect.ust.hk).}
	\thanks{$^{2}$Y. Yang is with School of Automation, Nanjing University of Information Science and Technology, Nanjing, China (e-mail: meyang@nuist.edu.cn). }
	\thanks{$^{3}$M. Y. Wang is with the Department of Mechanical
and Aerospace Engineering and the Department of Electronic and Computer
Engineering, Hong Kong University of Science and Technology, Hong Kong
(tel.: +852-34692544; e-mail: mywang@ust.hk).}%

}

\begin{document}

\maketitle
\thispagestyle{empty}
\pagestyle{empty}

\begin{abstract}

Monitoring the state of contact is essential for robotic devices, especially grippers that implement gecko-inspired adhesives where intimate contact is crucial for a firm attachment. However, due to the lack of deformable sensors, few have demonstrated tactile sensing for gecko grippers. We present Viko, an adaptive gecko gripper that utilizes vision-based tactile sensors to monitor contact state. The sensor provides high-resolution real-time measurements of contact area and shear force. Moreover, the sensor is adaptive, low-cost, and compact. We integrated gecko-inspired adhesives into the sensor surface without impeding its adaptiveness and performance. Using a robotic arm, we evaluate the performance of the gripper by a series of grasping test. The gripper has a maximum payload of 8$N$ even at a low fingertip pitch angle of 30$^\circ$. We also showcase the gripper’s ability to adjust fingertip pose for better contact using sensor feedback. Further, everyday object picking is presented as a demonstration of the gripper’s adaptiveness.

\end{abstract}


\section{Introduction}


In the past decade, gecko-inspired adhesives have been used to create robotic devices with unique functions, including climbing, braking, and grasping in microgravity \cite{Kim2008,Unver2009,Murphy2011,Kalouche2014,Hawkes2013,Hawkes2015b,Tse2020,Naclerio2019,Glick2018,Song2014,Dadkhah2016,Jiang2017,Roberge2018,Hawkes2016}. Gecko-inspired adhesives are advantageous over conventional attachment methods like pressure sensitive adhesives (PSAs) and suction cups due to its ability to switch swiftly between on and off without leaving any residues \cite{Parness2009,Hansen2005}. Moreover, gecko-inspired adhesives only require small preload forces to generate high adhesion at both normal and shear directions \cite{Autumn2006}. However, the strength of adhesion largely depends on the real contact area between the adhesives and the target surface, and loss of contact will cause attachment failure \cite{Autumn2000}. Therefore, it is crucial to develop a sensing system for gecko adhesives to monitor the contact area and force at the same time to avoid impending failure.


One prominent method to evaluate the contact state is to use vision-based tactile sensors. Previously, Eason et al. have successfully measured the stress distribution and contact area of a gecko toe using a high-resolution tactile sensor based on frustrated total internal reflection (FTIR) and a microtextured membrane \cite{Eason2015}. Contact information can be obtained by the change of brightness when the taxels are compressed. However, these sensors are too costly and bulky for integration into a compact robotic system such as a gripper. In other research, capacitive sensors were used to measure contact area and load conditions of gecko-inspired adhesives resting on thin film tiles \cite{Wu2015,Huh2018,Hashizume2019}. However, these methods impede the overall flexibility of the adhesives due to the non-stretchable nature of capacitive tactile sensors. When dealing with textured surfaces, the adhesives have the risk of failing due to low surface conformity. Moreover, measurement of stress distribution requires a densely packed array of capacitive sensors, which is sophisticated and costly. A compact contact state monitoring system for gecko gripper is still elusive. 

\begin{figure}
	\centering
	\includegraphics[width=0.485\textwidth]{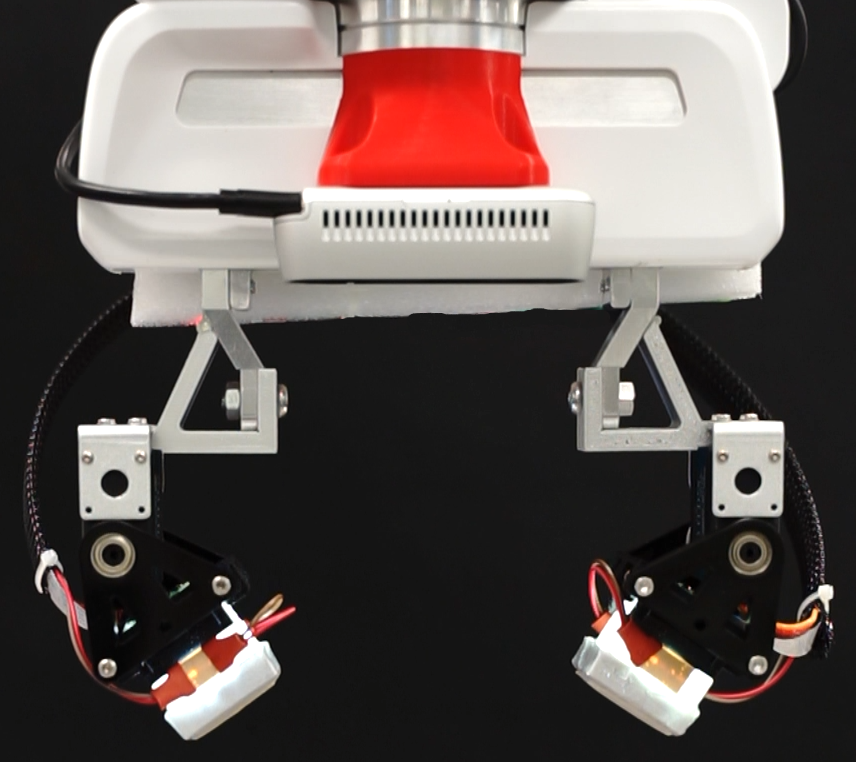}
	\vspace{-0.2cm}
	\caption{Dual-finger Viko gripper installed on Franka Emika Panda robotic arm.}
	\vspace{-0.4cm}
	\label{fig: intro_graph}
\end{figure}

Recently, vision-based tactile sensors have attracted the attention of many researchers due to their high resolution and sensitivity \cite{Dong2017,Yuan2015,Trueeb2020,Yamaguchi2017}. Unlike capacitive tactile sensors, vision-based tactile sensors measure contact information by converting contact signals into images. These sensors usually include a deformable elastomeric membrane with embedded markers and a camera. The adaptiveness of vision-based tactile sensors makes them preferable for contact measurements of gecko-inspired adhesives.  Previously, our group has successfully demonstrated slip detection using a custom-built, high-resolution vision-based tactile sensor called FingerVision \cite{Zhang2018}. The sensor is adaptive, compact, and low-cost.


Following this concept, we present Viko, an adaptive gecko gripper with vision-based tactile sensing capability, as shown in Fig. \ref{fig: intro_graph}. The gripper includes a modified version of our FingerVision sensor with denser markers and optical flow algorithms to measure contact area and shear force simultaneously at a higher resolution. Moreover, unlike previous capacitive sensors for gecko adhesives, the gecko-inspired adhesives are directly integrated onto the deformable membrane surface of our vision-based tactile
sensor. Therefore, the gripper can maintain its adaptiveness to sustain higher loads, even on curved or textured surfaces. The main contribution of this work can be differentiated as follows: 
\begin{enumerate}
\item Development of a compact and high-yield vision-based tactile sensor capable of monitoring contact area and shear force simultaneously at high resolution. 
\item Direct integration of gecko-inspired adhesives to the gripper without impeding the overall adaptiveness. 
\item Demonstration of readjusting fingertip pose based on sensor feedback for better attachment.  
\end{enumerate}


\section{Design and Fabrication}

\begin{figure}
	\centering
	\includegraphics[width=0.49\textwidth]{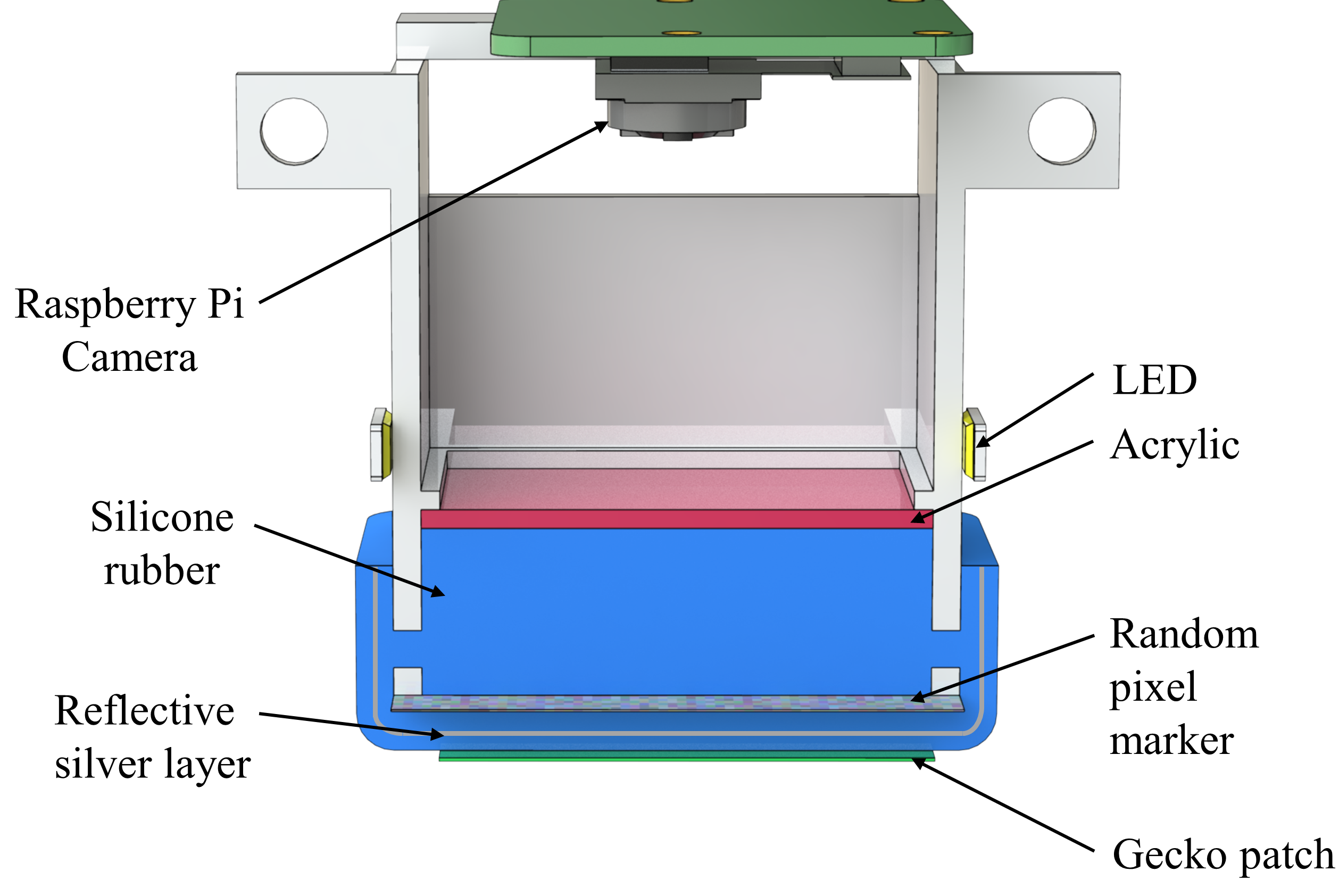}
	\vspace{-0.2cm}
	\caption{Cross-section view and decomposition of the vision-based tactile sensor.}
	\vspace{-0.4cm}
	\label{fig: Sensor_illu}
\end{figure}

\subsection{Adaptive Gecko Gripper}
The adaptive gecko gripper includes two opposing fingertips mounted on a commercially available parallel gripper produced by Franka Emika, as shown in Fig. \ref{fig: intro_graph}. The fingertips and the parallel gripper are connected via servo motors and 3D printed frames. The servo motors allow the fingertips to rotate for better alignment to the target object. The detailed composition of each fingertip depicted, as shown in Fig. \ref{fig: Sensor_illu}, is composed of a custom-built vision-based tactile sensor covered with gecko-inspired adhesives. The adhesives can be directly bonded to the elastomeric membrane of the sensor by double-sided tape or casting without impeding the gripper’s passive adaptiveness and sensor performance.

The gripper approaches the target object with an approximate fingertip pose to make the first grasping attempt. As the fingertips contact the object, the state of contact would be determined by the contact area feedback from the sensor. Subsequently, fingertip orientation can be adjusted by the servo motor to align with the target surface for the maximum contact area. For objects with curved or textured surfaces, the deformable elastomeric membrane on the sensor can conform to make good contact. Once the fingertip orientation is determined, we apply a preload to the fingertips by the parallel gripper. The microwedge structures on the gecko-inspired adhesives bend when loaded in shear and generate adhesion.


\begin{figure}
	\centering
	\includegraphics[width=0.485\textwidth]{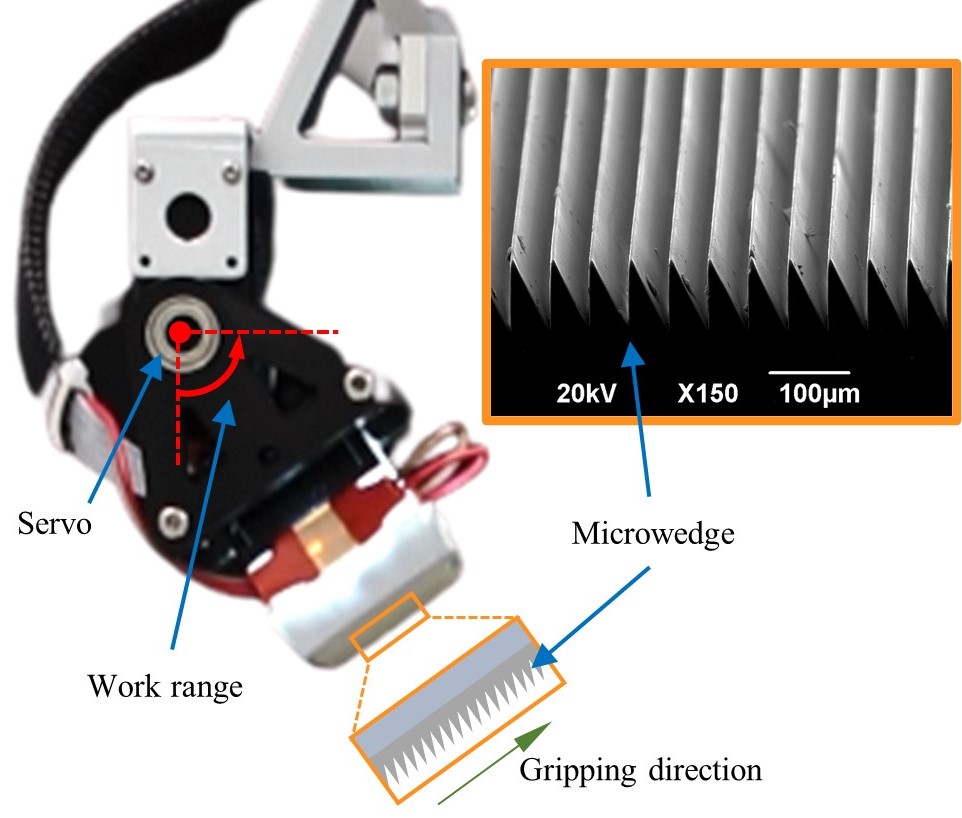}
	\vspace{-0.2cm}
	\caption{Detailed view of the gripper unit showing the gecko-inspired adhesive layer with a zoomed image under scanning electron microscope (SEM). The schematic illustrates the geometry of the microwedge structure and the gripping direction.}
	\vspace{-0.4cm}
	\label{fig: gripper_illu}
\end{figure}

\begin{figure}
	\centering
	\includegraphics[width=0.485\textwidth]{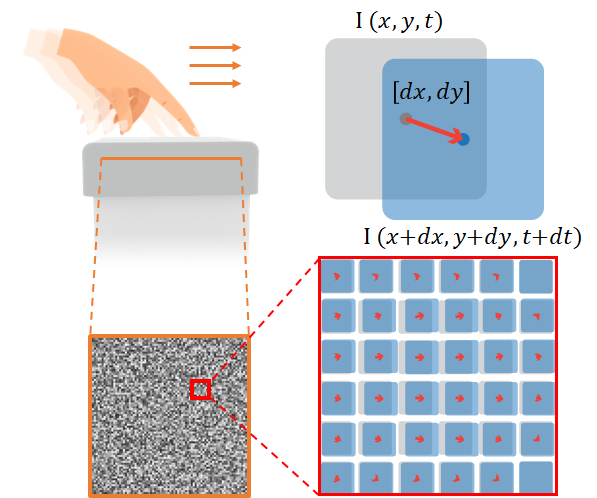}
	\vspace{-0.2cm}
	\caption{Illustration of the working principle of the vision-based tactile sensor. (Top, left) Illustration of a fingertip sliding on the sensor. (Bottom, left) The sensor captures the grayscale image. (Bottom, right) Zoom-in sketch of the grayscale image in which the gray squares represent the initial pixel location$(x,y)$, the blue squares represent current pixel location $(x\text{+}dx,y\text{+}dy)$, and the red arrow is the pixel translation vector. (Top, right) A single-pixel translation from time $t$ to $t+dt$.}
	\vspace{-0.5cm}
	\label{fig: optical_flow}
\end{figure}

\subsection{Vision-based Tactile Sensor}

For measurement of the gripper’s contact state during grasping, we redesigned our FingerVision tactile sensor. Instead of slip detection, the focus of this paper is to measure the contact area and shear force , which are essential for gecko-inspired adhesive. The mainframe of the sensor is 3D printed with a Stereolithography (SLA) printer using an opaque white resin to reduce the light interference from the ambient environment. For a bright and uniform illumination, we used two symmetrically arranged white LED light sources mounted outside the mainframe. The light transmits through the resin, which results in large and even diffusion. 


The deformable elastomeric membrane resting on a translucent acrylic plate contains several layers, and it is fabricated on a layer by layer basis. We used a very soft and translucent silicone rubber (Solaris, Smooth-On, USA) as the primary casting material to facilitate high adaptiveness. To achieve high sensitivity and spatial resolution to contact force, we first embedded a layer of dense random pixel marker into the elastomeric membrane as the tracking feature. Afterward, a thin reflective silver layer is sprayed evenly to the surface of the previous layer for uniform illumination inside the sensor and prevents the penetration of exterior lightings. Finally, another layer of silicone rubber is added on top of the silver layer for protection. 

The displacements of the markers are acquired by a commercially available Raspberry Pi camera mounted on the top of the sensor. The focal length of the camera we use is 3.04$mm$, so that the camera is mounted 30$mm$ away from the markers for a 35x35$mm$ sensing area. Since the overall size of the sensor is limited by the camera focal length and field of view, we can also adopt a fisheye camera to make the sensor thinner and more compact. However, due to image distortions, an extra calibration process is required.

The dense inverse search (DIS) optical flow algorithm \cite{Gevers2016} was applied to evaluate the translation of dense random markers captured by the camera. This algorithm was firstly applied in vision-based tactile sensing in \cite{Sferrazza2019}. Instead of processing with the RGB colored image, the intensity variation in the grayscale image is enough for differentiating individual markers while lower the run-time. The deformation of the elastic membrane occurs when the external force stimulation is applied, as illustrated in Fig. \ref{fig: optical_flow}, the markers will be distorted correspondingly. Hence, the pixel displacement can be traced by the optical flow algorithm based on consecutive image flows of the markers up to 40 frames per second (FPS). The translation of each pixel can be correctly traced without mismatching to the adjacent pixel due to the random pixel pattern and high FPS. Random pixel pattern prevents the same grayscale intensity between adjacent pixels, while high FPS lowers the derivation between frames. By comparing the current and initial pixel location, a dense pixel translation vector field can be deduced to obtain the contact area and shear force. Afterward, the contact area is acquired by computing the divergence of the pixel translation vector field filtered by thresholding with fine-tuned value to eliminate noise. Shear force can also be calculated from the pixel translation vector field, which will be further discussed in Section. \Romannum{3}, part A. 

\subsection{Gecko-inspired Adhesives}
As shown in Fig. \ref{fig: gripper_illu}, the surface of the gecko-inspired adhesives contains wedge-shaped microstructures, which have a height of 100$\mu m$ and a sharp tip angle of 25.5$^\circ$. For each fingertip, an adhesive pad of 30x30$mm$ is used. We create the adhesives using silicone rubber (Sylgard 184, Dow Corning) by a molding method previously described in \cite{Tse2020}. As the adhesives engaged the target surface and loaded in shear direction, the wedge-shaped microstructures bend to increase the contact area, thus generating large adhesion.

\section{Sensor Calibration and Validation}

\subsection{Shear Force Calibration}
\begin{figure*}
	\centering
	\includegraphics[width=0.99\textwidth]{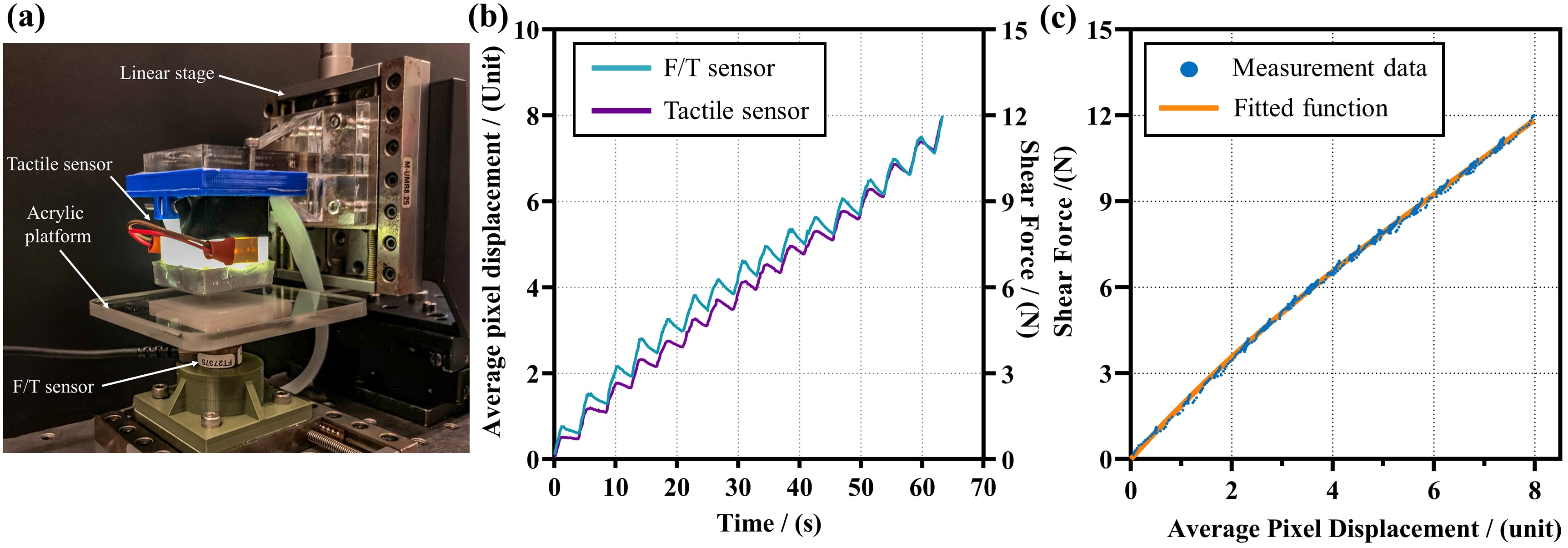}
	\caption{(a). Experiment setup of shear force calibration. (b). Comparison of shear force measurements from the vision-based tactile sensor (left axis) and F/T sensor (right axis). (c). The calibration result of average pixel displacement to shear force, where blue dots represents the measurement data, and the orange curve represents the fitted cubic polynomial function given by Eq. \ref{shear_map}.}
	\label{fig:fv_combine}
\end{figure*}

To calibrate the sensor, we acquired data through a custom-built platform with a high precision 6-axis ATI Nano-17 force/torque sensor(F/T sensor, resolution: 0.12$N$, range: $\pm$12$N$) and a 3-axis manual linear stage (NEWPORT M-UMR8.25) as shown in Fig. \ref{fig:fv_combine}(a). For a more consistent analysis of elastic materials like silicone rubber, some assumptions were made to fit our application as follows:
\begin{enumerate}
\item The rigid boundary constraints of the sensor frame are neglected; thus, the displacement of each pixel is assumed to be consistent, given the same loading condition.
\item The resized sensing surface (22x22$mm$) of the elastic membrane is assumed to be flat. 
\end{enumerate}

Based on these assumptions, the sensing area was resized to 22x22$mm$ of the center part, which could lower the effect of boundary conditions and surface non-flatness.

The vision-based tactile sensor was first brought into contact with the acrylic plate, which is mounted on the F/T sensor at a constant load of 5$N$. Displacement in shear direction was applied to the tactile sensor manually along the gripping direction in 16 steps, each with a constant 0.25$mm$ increment. Shear force data were recorded at 125$Hz$ by ATI data acquisition software (ATIDAQ) simultaneously with the optical flow from the tactile sensor at 25$Hz$. The acquired shear force data was processed by MATLAB before calibration. To match the sampling frequency between the two sensors, we sampled the shear force data from the F/T sensor again at 25$Hz$ and synchronized the timeline. For the calculation of the average pixel displacements, we divided the summation of the optical flow data by the total pixel number. The results are plotted in Fig. \ref{fig:fv_combine}(b), where we observed small shear force reductions in each step. This can be explained by a combined effect of the viscoelasticity of the elastomeric membrane and slipping in the contact surface. The tactile sensor data also reflects the same phenomenon with negligible phase variation, which indicates high agreement with the F/T sensor. Afterward, we adopted a cubic polynomial function to fit the shear force data with the average pixel displacement, with $R^2 =$ 0.999, as plotted in Fig. \ref{fig:fv_combine}(c). The shear force, $\tau$ can then be mapped as: 

\begin{equation}
    \tau (x)=1.966x-0.1033x^2+0.005353x^3
    \label{shear_map}
\end{equation}

where $x$ denotes the average pixel displacement.




\subsection{Contact Area Validation}

\begin{figure}
	\centering
	\includegraphics[width=0.485\textwidth]{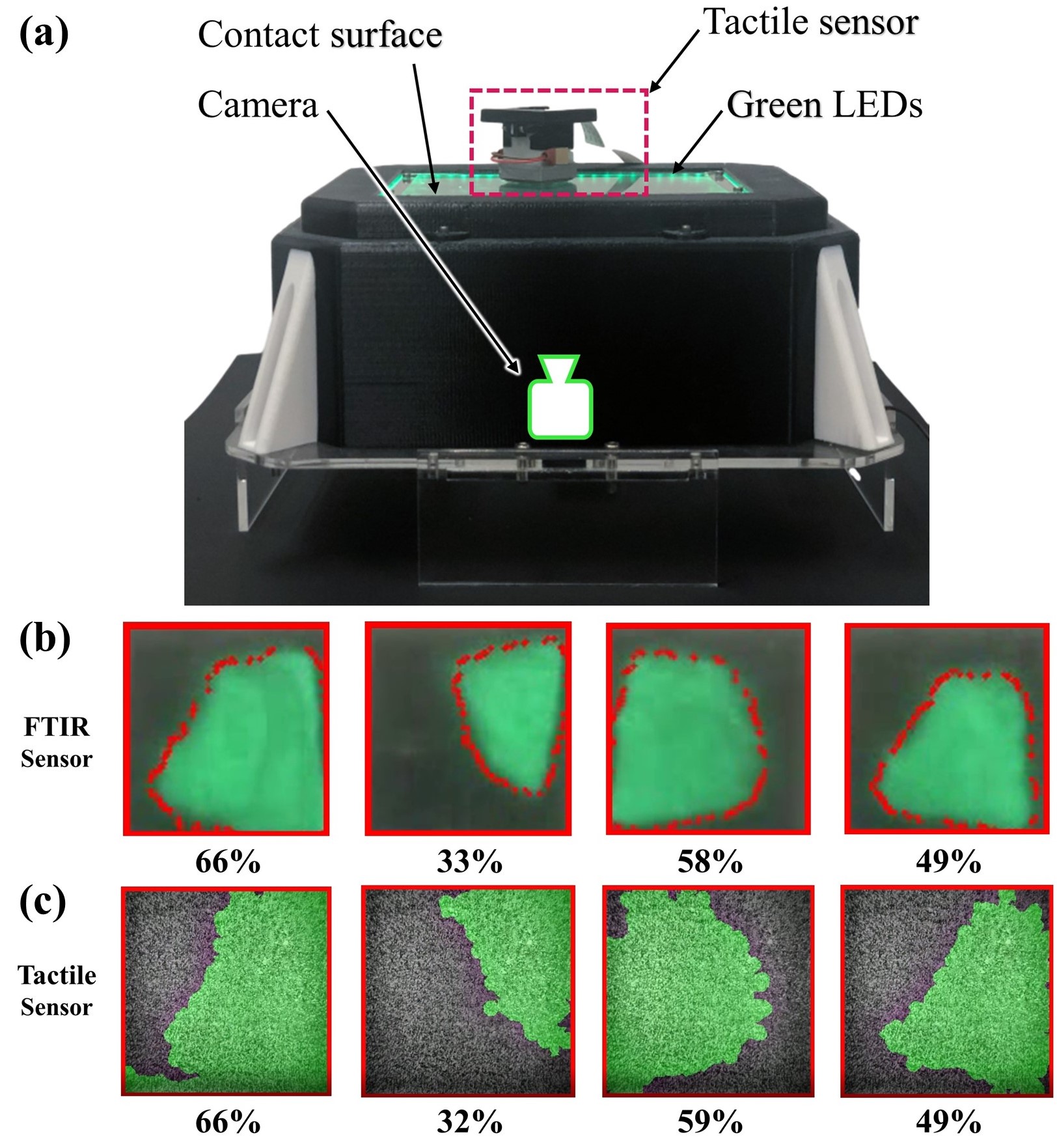}
	\caption{(a). Contact measurement platform based on FTIR sensor, with green LED illumination on the edge of acrylic surface. A wide-angle camera marked with  green outline is positioned at the center beneath the contact surface for image capture. Contact area measurement comparison between FTIR (b) and tactile sensor (c), the percentage of contact are listed for comparison. }
	\label{fig:ft_combine}
\end{figure}

FTIR sensor has been successfully adopted by many groups as a validation tool for contact area measurements using tactile sensors \cite{Hawkes2015}. Therefore, we validate the contact area measurement results from the vision-based tactile sensor via a custom-built flat FTIR platform with evenly distributed green LEDs illuminating the boundary of an acrylic surface, as shown in Fig. \ref{fig:ft_combine}(a). The contact surface image is captured by a commercially available USB camera with 25 FPS. Varying the orientation of the tactile sensor while keeping contact with the FTIR acrylic surface, the contact region followed the changes as shown in the supporting video. In Fig. \ref{fig:ft_combine}(b,c), the contact regions sensed by both sensors are represented by the bright green area. Since the FTIR sensor gives a diffusive boundary, the contact region is marked with red dots along the peripheral for identification. By comparing the measurement results, we found a good agreement between the vision-based tactile sensor and the FTIR platform in terms of the percentage of contact and peripheral geometry. 

\section{Experiments and Results}
\begin{figure*}[h]
	\centering
	\includegraphics[width=0.99\textwidth]{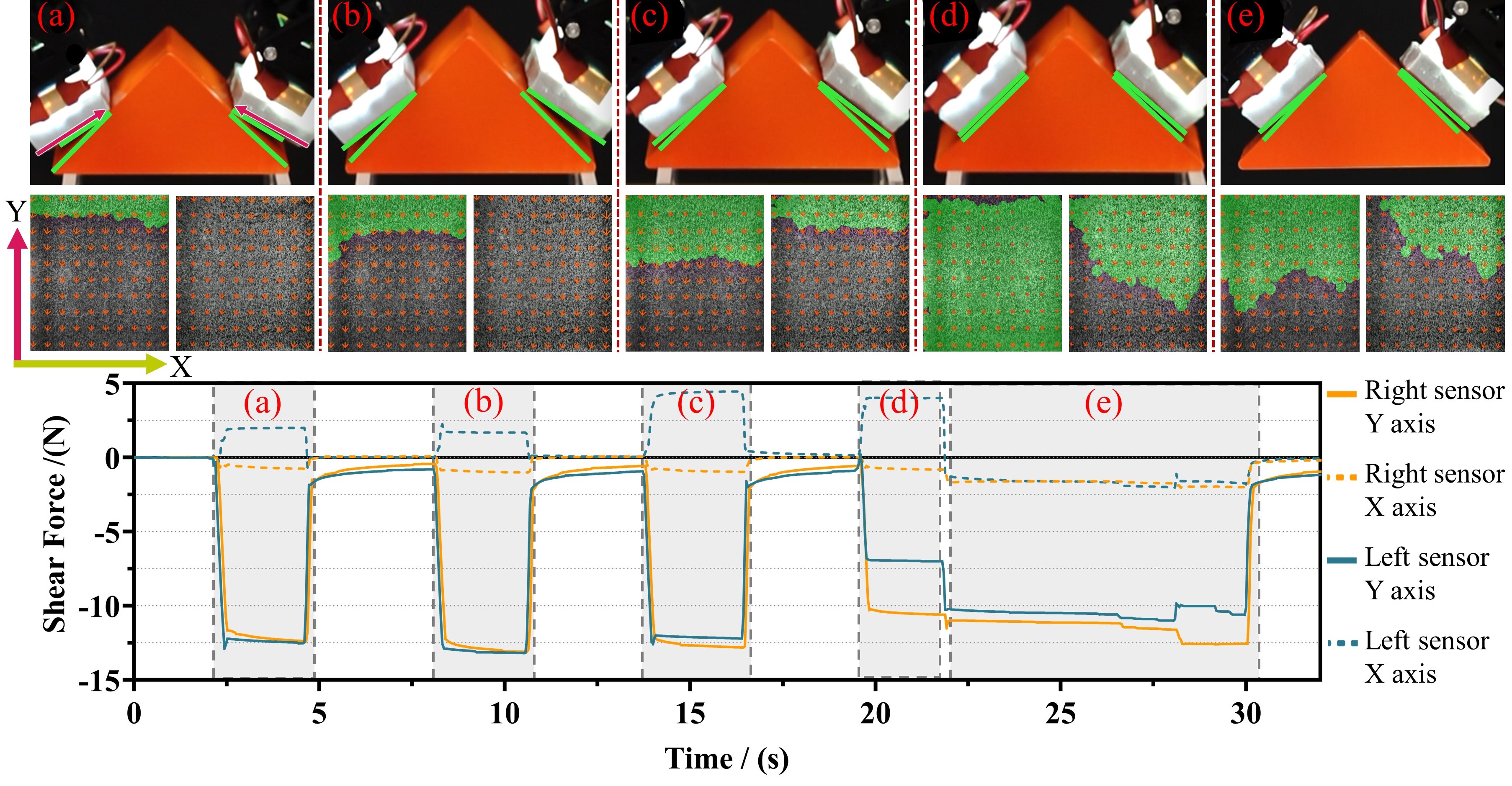}
	\caption{(Top) Figures demonstrate the contact condition with a various fingertip pitch angles of grasping(a-d) and pick up(e). (Middle) The corresponding tactile sensor measurements of contact area(light green area) and shear force(orange vector). (Bottom). Detailed shear force results of the left and right tactile sensor.}
	\label{fig: Exp_22}
\end{figure*}


Several experiments were designed to evaluate the performance of the Viko gripper. This section can be divided into two main parts. In the first part, we validated the gripper's ability to readjust fingertip pose based on the contact information from sensor feedback under a real grasping scenario. In the second part, we demonstrated the grasping of everyday object with various geometries, substantiating the adaptiveness and robustness of the gripper. Notably, the gripper is able to handle oversized objects, which is difficult for conventional grippers without gecko-inspired adhesives. 


Raspberry Pi (Model 3B+) was used as the hardware interface for data transmission between the computer and the Viko gripper via ethernet cable. We used robot operating system (ROS) network on a computer as the master for receiving camera data and publishing control signals to the servo motors and robotic arm. All computations, including image processing, gripper planning, and robotic arm controlling, were done by the computer.

\subsection{Sensor-based Grasping}

In practical applications, firm attachment of gecko-inspired adhesives requires good contact with the target surface. Misalignments between the adhesives and the target surface may lead to poor contact and grasping failure. Therefore, a series of tests were conducted to evaluate the sensor performance under real grasping scenarios and demonstrate the gripper's ability to readjust fingertip pose for better contact based on sensor feedback.

We used an isosceles right triangular block as the grasping object and engaged the gripper at a random initial fingertip pitch angle, as shown in Fig. \ref{fig: Exp_22}(top). In the beginning, the contact angle between the adhesives and the target surface was different on each fingertip. For the first two approaches, only a small portion on the top of the left fingertip was in contact, as shown in Fig. \ref{fig: Exp_22}(a, b). This can also be reflected by the contact area readings marked with light green from the sensors. The right fingertip had no measurable contact area, and the grasping would most likely fail. Arrows in Fig. \ref{fig: Exp_22}(middle) shows the magnitude and direction of the shear force. Realtime measurement of total shear force in x-axis and y-axis are plotted in Fig. \ref{fig: Exp_22}(bottom). Although no contact area was measured, compression at the edge of the elastomeric membrane caused deformation. As a result, shear force was measured. In the next two attempts, the servo motor rotated to align the adhesives and the target surface for a larger contact area. Observation of the sensor readings showed an increase in the contact area and was in agreement with the actual state. Finally, the gripper picked up the object after good contact was achieved, and the contact area readings in two fingertips showed a tendency to achieve balance, as shown in Fig. \ref{fig: Exp_22}(e).  Placement of the object can be shown by the slight fluctuation in total shear force before detachment.

\begin{figure}
	\centering
	\includegraphics[width=0.495\textwidth]{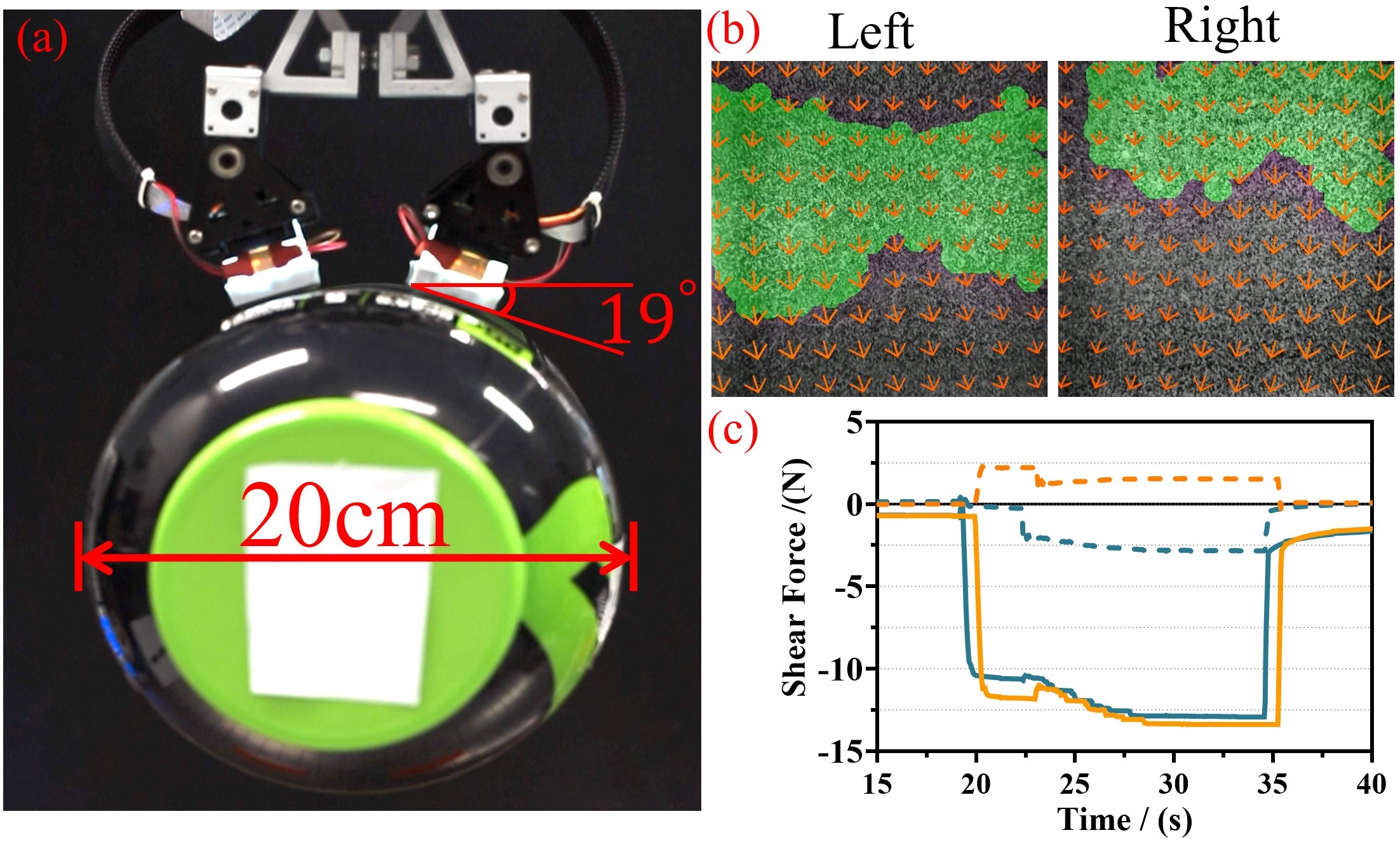}
	\vspace{-0.2cm}
	\caption{(a). Viko gripper grasping an oversized cylindrical bottle. (b). Corresponding tactile sensor results under the same coordinate system labeled in Fig. \ref{fig: Exp_22}. (c). The shear force measurement plot.}
	\vspace{-0.4cm}
	\label{fig: Exp_14}
\end{figure}

Results of the experiment indicate that vision-based tactile sensors can provide realtime contact state monitoring during grasping operations, making a firm base for robotic manipulation control. Such information may help the gripper to achieve better grasping performance. 


\subsection{Adaptive Grasping}

To evaluate the gripper's adaptiveness to object variations. We measured the gripper's maximum load by testing it on objects of varying size, shape, surface texture, and stiffness.

For maximum loading test, three isosceles triangular blocks with a base angle of $60^\circ$, $45^\circ$, and $30^\circ$ were used to examine the maximum gripper payload under various fingertip pitch angles. The testing blocks were placed on a platform, where we adjusted the gripper to align the adhesives with the block surfaces and applied constant preload. A cable tie was used to connect the testing block to a force gauge (AIGU NK-100). Subsequently, the force gauge was pulled downward, perpendicular to the testing block until the gripper and testing block detached. The test was repeated three times for a more consistent result. The recorded average maximum loads are 18, 11, and 8 $N$ with respect to $60^\circ$, $45^\circ$, and $30^\circ$ fingertip pitching angle. The results indicate that the performance drops as the fingertip pitch angle decreases. This can be explained by the weaker normal adhesion of gecko-inspired adhesives compared to its shear adhesion. However, the loading is still considerably high when grasping at low fingertip pitch angles in comparison with grippers without gecko-inspired adhesives.

\begin{figure}
	\centering
	\includegraphics[width=0.49\textwidth]{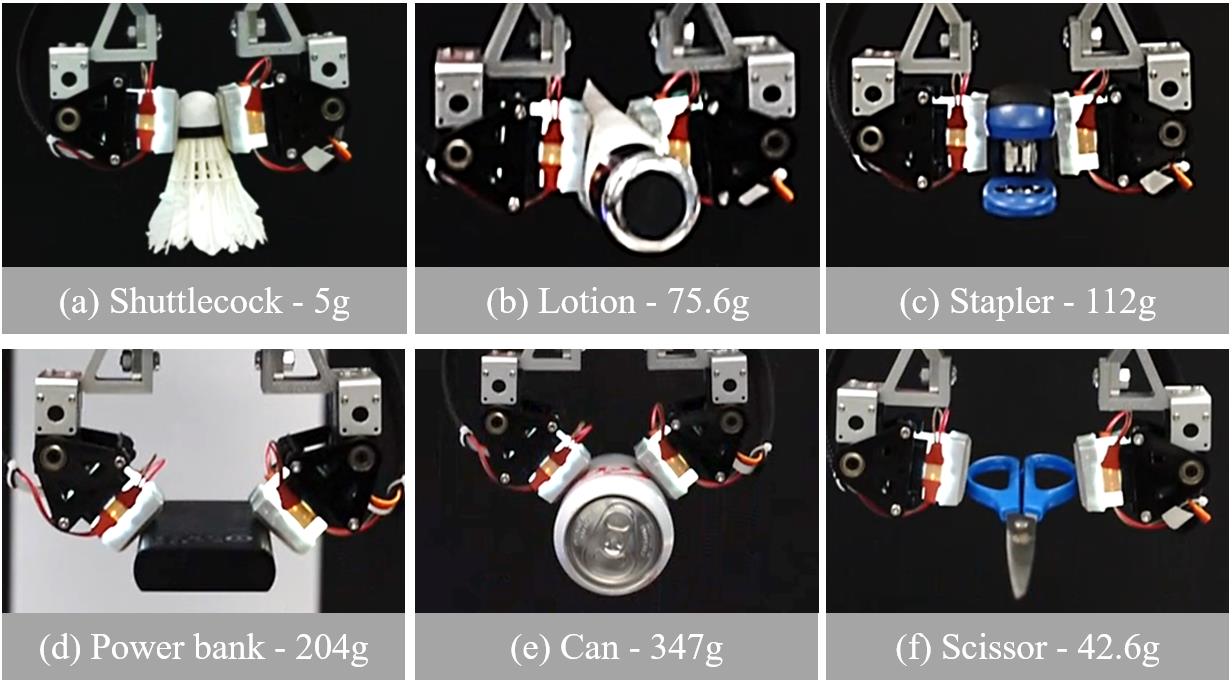}
	\vspace{-0.2cm}
	\caption{Viko gripper grasping various everyday objects.}
	\vspace{-0.2cm}
	\label{fig: Exp_multi}
\end{figure}

To test the gripper performance on oversized objects, we used a plastic bottle (263$g$) with a diameter of 20$cm$, which is larger than the distance between two parallel fingertips (10$cm$). The gripper performed pick and place motion with a $19^\circ$ fingertip pitch angle, as illustrated in Fig. \ref{fig: Exp_14}. The elastomeric membrane conformed to the shape of the bottle to enhance the contact area. This phenomenon is validated by the sensor readings, which shows that the contact area is around $40-50\%$ during the grasping process, as plotted in Fig. \ref{fig: Exp_14}(b). The shear force diagram plotted in Fig. \ref{fig: Exp_14}(c), also demonstrates clear boundaries that identify the moment of contact and detachment. Moreover, a high shear force was maintained at around 10 to 12$N$ throughout the motion, which implies that the gripper performed a firm grasp. The ability to provide high grasping force at low fingertip pitch angles enables the gripper to firmly handle oversized objects, which is not possible for most conventional grippers. This characteristic makes it advantageous in operating large objects within a constrained space where large grippers cannot be deployed.



The adaptiveness of the gripper is further evaluated by testing on several everyday objects, as demonstrated in Fig. \ref{fig: Exp_multi}. We followed the setup in the maximum loading tests and performed pick and place motion. Before testing, we adjusted the fingertip pose according to the object shape and local geometry. For objects with a curved edge (d, e), the gripper can reduce the fingertip pitch angle; while dealing with objects with a smaller size and rigid body (a, c, f), the gripper can be tuned to parallel mode to maximized shear force; and it can also grasp the soft body object (b). 

\section{Conclusion and Future Works}


In this research, we present the design of an adaptive gecko gripper with a vision-based tactile sensor (Viko). A high-resolution vision-based tactile sensor is developed to the realtime monitor contact area and shear force for gecko-inspired adhesives. The adhesives can be integrated into the sensor surface while maintaining its high adaptiveness and performance. The shear force and the contact area measurement of the sensor are validated by an F/T sensor and FTIR optical imaging, respectively. Finally, we show the gripper’s ability to improve contact based on sensor feedback and evaluate the gripper’s performance on a variety of objects with different surface geometries via grasping tests. Viko offers a new angle to designing a controlled robotic system that utilizes gecko-inspired adhesives. 


For future work, we would continue to improve the performance of the sensor in terms of accuracy and sampling frequency. Also, we plan to include sensing of normal force and depth information for better contact event prediction. These improvements may lead to a more intelligent robotic system with closed-loop control. Applications based on the idea of this paper can also be extended to other robotic devices like gecko climbing robots.


\bibliographystyle{IEEEtran}
\bibliography{reference}

\end{document}